\documentclass{IOS-Book-Article}     
\usepackage{graphicx}
\usepackage{booktabs}       
\usepackage{verbatim}
\newcommand{\wrt}{{\it w.r.t. }}   
\newcommand{\ie}{\emph{i.e.}, }      

\usepackage{subfiles}
\usepackage{amsmath}
\usepackage{subcaption}

\begin{document}
\begin{frontmatter}          
%
\title{Feature discriminativity estimation in CNNs for transfer learning}
\runningtitle{}

%

\author[B]{\fnms{Victor} \snm{Gimenez-Abalos}}, %
\author[A]{\fnms{Armand} \snm{Vilalta}}, %
\author[A]{\fnms{Dario} \snm{Garcia-Gasulla}}, %
\author[B]{\fnms{Jesus} \snm{Labarta}}, %
and
\author[B]{\fnms{Eduard} \snm{Ayguad\'{e}}}
\address[A]{Barcelona Supercomputing Center (BSC)\\ (\{armand.vilalta, dario.garcia\}@bsc.es)}
\address[B]{Universitat Polit\`ecnica de Catalunya - BarcelonaTech (UPC)\\ (victor.gimenez.abalos@est.fib.upc.edu)}

\begin{abstract}
The purpose of feature extraction on convolutional neural networks is to reuse deep representations learnt for a pre-trained model to solve a new, potentially unrelated problem. However, raw feature extraction from all layers is unfeasible given the massive size of these networks. Recently, a supervised method using complexity reduction was proposed, resulting in significant improvements in performance for transfer learning tasks. This approach first computes the discriminative power of features, and then discretises them using thresholds computed for the task. In this paper, we analyse the behaviour of these thresholds, with the purpose of finding a methodology for their estimation. After a comprehensive study, we find a very strong correlation between problem size and threshold value, with coefficient of determination above 90\%. These results allow us to propose a unified model for threshold estimation, with potential application to transfer learning tasks.

\end{abstract}


\end{frontmatter}


\section{Introduction}

Convolutional Neural Networks (CNNs) have become the new standard approach for dealing with image processing tasks. These models require exhaustive and expensive training processes, which result in particularly rich and useful representations \cite{lecun2015deep,yosinski2014transferable,carter2019activation}. Unfortunately, these methods have strong requirements regarding dataset size, computational power and expert optimisation. For any task in which these factors are an issue, training the model from scratch becomes unfeasible.

Transfer learning studies how to extract and reuse the representations encoded within pre-trained deep neural networks. Among other things, through transfer learning one can exploit deep representations through alternative machine learning methods, without having to train a deep net \cite{azizpour2016factors,sharif2014cnn,gong2014multi}. This is also known as \textit{transfer learning for feature extraction}.

Transfer learning for feature extraction is based on performing feed-forward passes through a pre-trained neural network (trained on the \textbf{source task}), while feeding it data instances of a new task (the \textbf{target task}). In CNNs, this is often done to transform the data from the image domain into a numerical, high-dimensional space potentially suitable for a wide variety of machine learning methods. This representation, also known as an \textit{embedding}, is typically composed by a subset of internal neural activations. This feature extraction process is independent of dataset size and relatively cheap in computational terms, since it only implies feed-forward passes through the CNN. For the same reason, it requires no hyper-parameter optimisation.

A key factor in the performance of transfer learning solution is the selection of features to extract, and their postprocessing. Given the size of most CNNs, the number of raw features can easily be in the thousands, making it challenging to most machine learning methods due to the curse of dimensionality \cite{hughes1968mean}. To mitigate that, one can reduce the embedding dimensionality by removing features. However this approach implies a vocabulary loss, which is not desirable. An alternative is to reduce the embedding space instead, for example through discretisation \cite{garcia2018out}. 

To decide which features to remove or how to discretise the space, it is useful to have a pre-computed measure of feature discriminativity on the target task. This can be done supervisedly, through Kolmogorov-Smirnov distances. In previous work \cite{garcia2018behavior} these distances were discretised using two thresholds (one for discriminative feature activation and one for inhibition). Unfortunately, as this method implies recomputing the discriminativity after a random shuffle, it is costly and stochastic.

The goal of this paper is to find an alternative method to discretise the discriminativity space into expected feature behaviour in presence of a class. We base our method on an existing correlation between the average number of instances per class and optimal threshold value. This approach has no stochasticity and reduces the computational cost of the original method. We show how more than 90\% of the empirical thresholds' variability is explained by the average instances per class of the target task. Our results indicate that the remaining 10\% is strongly correlated with the class imbalance of the target task, and its similarity to the source task.

\section{Methods}

We focus on the relation between the average number of instances per class in the target task, $\hat{I_c}$, and the optimal feature discriminativity thresholds ($t^+$ and $t^-$). $\hat{I_c}$ values for all the tasks considered are shown in Table \ref{tab:datasets}. How we obtain discriminativity thresholds is described next.

To determine feature discriminativity for a target task, we first extract neural activations from the pre-trained model. The first processing done to this raw embedding, is an average spatial pooling on the convolutional filters (to disregard activation location). Afterwards, we perform a feature-wise standardisation across all instances of the target task (\ie we compute feature-wise z-scores). This results in a standardised embedding.

For each feature $f$ in the standardised embedding, and each class label $c$ in the target task, we define $D_{KS}(f,c)$. This is computed as the signed Kolmogorov-Smirnov (KS) distance between the activations of $f$ on the instances of $c$ (inner-class), and on the rest (outer-class). $D_{KS}(f,c)$ are values between -1 and 1 indicating the peculiarity of $f$ behaviour for $c$, either by over-activation or under-activation. We will also refer to this as the \textbf{feature-class pair discriminativity}.

The relevant $D_{KS}(f,c)$ values are somewhere between zero (arbitrary feature behaviour) and -1 or 1 (unique behaviour). To maximise the trade-off between noise and information, previous work compared the $D_{KS}$ values obtained for a given class, with those obtained by a set of instances randomly chosen from all target task classes \cite{garcia2018behavior}. These values represent the discriminativity value of noise. A second KS test between these two distributions of $D_{KS}$ values provides the threshold that maximises the distance between information and noise. This is done independently for positive and negative $D_{KS}$ values giving rise to $t^+$ and $t^-$. The thresholds may be used for classifying the feature-class pairs in the previous work to perform embedding discretisation\cite{garcia2018out} or for dimensionality reduction.

\subsection{Proposed methodology}

To avoid the short-comings of the shuffling methodology for obtaining the thresholds, namely the stochasticity and computational complexity, we propose to find a regression model fitting their behaviour based on the target task $\hat{I_c}$. We fit and evaluate such regression using the empirical approach of the shuffling method using several target tasks.

Since we are using the shuffling methodology to guide the regression fitting, we first must make sure that such shuffling results in stable thresholds. For that purpose, we repeat the random shuffling 21 times, resulting in 21 sets of thresholds (see \ref{exp:stoch} for further details). We measure the stability of the shuffling methodology through the standard deviation of these values.


Previous results \cite{garcia2018behavior} find high correlation between $\hat{I_c}$ and optimal feature discriminativity thresholds. Analysis of the methodology hints to the presence of a horizontal asymptote as $\hat{I_c}$ increases. At the same time, the absolute value of the thresholds seems to be inversely correlated to $\hat{I_c}$. For this reason, we discard the use of a linear regression. In preliminary studies we considered the following alternatives: the logarithmic, reciprocal and logarithmic reciprocal. The results obtained by the logarithmic reciprocal are remarkably better than the alternatives, which is why these are the only results we show and discuss in the rest of this work. Formally, the logarithmic reciprocal is as follows:

\begin{equation}
t(\hat{I_c};a,b)=a+b/ln(\hat{I_c})
\label{eq:log_recip}
\end{equation}

\section{Experiments}

Our goal is to find a versatile model for threshold estimation. For this purpose, we study our regression model using two CNN architectures, two source tasks and twenty-one target tasks, since this is the most influential component.

For the CNN architectures we use the VGG16 and VGG19 topologies\cite{simonyan2014very}. These are composed by consecutive blocks of convolution and pooling layers (16 and 19 layers respectively), and two fully connected layers. This sort of architecture is quite representative of the CNN designs being used today. As for the source tasks, we use the following: \textit{ImageNet 2012}\cite{russakovsky2015imagenet}, a dataset for classification spanning 1000 categories of objects, and \textit{Places 2}\cite{zhou2016places}, a scene recognition task unrelated to \textit{ImageNet 2012} with less categories. Of the possible combinations of architecture-source task, the only case we do not have available is the VGG19 trained on \textit{Places2}. The rest are referenced as follows: VGG16 CNN trained on \textit{ImageNet 2012} (\textit{VGG16IN}), VGG19 CNN trained on \textit{ImageNet 2012} (\textit{VGG19IN}), and VGG16 CNN trained on \textit{Places2} (\textit{VGG16P2}).

\subsection{Target tasks}

Since we want to obtain a generalisable method, we need to use different target tasks, ideally with different $\hat{I_c})$ and spanning different domains. We consider the following 10 datasets, freely available online:

\begin{itemize}
    \item \textit{MIT Indoor Scene Recognition} dataset \cite{quattoni2009recognizing} (\textit{mit67})
    \item \textit{Caltech-UCSD Birds-200-2011} dataset \cite{wah2011caltech} (\textit{cub200})
    \item \textit{Oxford Flower} dataset \cite{nilsback2008automated} (\textit{flowers102})
    \item \textit{Oxford-IIIT-Pet} dataset \cite{parkhi2012cats} (\textit{cats-dogs})
    \item \textit{Stanford Dogs} dataset \cite{khosla2011novel} (\textit{stanforddogs})
    \item \textit{Caltech 101} dataset \cite{fei2007learning} (\textit{caltech101})
    \item \textit{Caltech 256} dataset \cite{fei2007learning} (\textit{caltech101})
    \item \textit{Food-101} dataset \cite{bossard2014food} (\textit{food101})
    \item \textit{Describable Textures} Dataset \cite{cimpoi2014describing} (\textit{textures})
    \item \textit{Oulu Knots} dataset \cite{silven2003wood} (\textit{wood})
\end{itemize}

To increase the number of target tasks feeding our regression, while providing variance in $\hat{I_c}$, in some cases we consider the different data splits originally provided as different tasks. In particular, we use training sets (TR), test sets (TE), joined training and test sets (TRTE) and validation sets (VAL). From now on, all references to target tasks will regard to a specific dataset and split. The properties of the 21 resulting target tasks are shown in Table \ref{tab:datasets}. Notice the \textit{caltech101TRTE} (followed by \textit{caltech256TRTE}) has a remarkably larger imbalance in the number of instances per class than the rest of tasks.

\begin{table}[tbp!]
    \caption{Properties of all tasks used in our experiments, including average number of instances per class ($\hat{I_c}$ ) and the corresponding standard deviation (Imbalance).}
    \label{tab:datasets}
    \centering
    \scalebox{1}{
    \begin{tabular}{lrrrrr}
        \toprule
        Target task/s    & \#Images  & \#Classes    & $\hat{I_c}$ & Imbalance\\
        \midrule
caltech101TRTE & 9145 & 102 & 90 & 123.07\\
caltech256TRTE & 30607 & 257 & 119 & 85.69\\
catsdogsTR/TE & 3669/3680 & 37 & 99/99 & 1.5/1.5\\
cub200TR/TE/TRTE & 5994/5794/11788 & 200 & 30/29/59 & 0.17/2.91/2.91\\
flowers102TR/VAL/TE & 1020/1020/6149 & 102 & 10/10/60 & 0/0/44\\
food101TE & 25250 & 101 & 250 & 0\\
mit67TR/TE/TRTE & 5360/1340/6700 & 67 & 80/20/100 & 1.39/1.39/0\\
stanforddogsTR/TE & 12000/8580 & 120 & 100/72 & 0/23.12\\
texturesTR/VAL/TE & 1880/1880/1880 & 47 & 40/40/40 & 0/0/0\\
woodTR & 438 & 7 & 62 & 50.84\\
        \bottomrule
    \end{tabular}
    }
\end{table}

\subsection{Method stochasticity with respect to target dataset} \label{exp:stoch}
As previously discussed, we need to assess the stability of the shuffling methodology, since we will be using it to validate the consistency of our regression model. Due to computational constraints, we only use a subset of target tasks: \textit{caltech101TRTE}, \textit{mit67TRTE}, \textit{cub200TR}, and \textit{flowers102TR}. This subset spans different topics, have different $\hat{I_c}$, and different imbalance levels. We use the pre-trained model \textit{VGG16IN} to obtain the corresponding random $D_{KS}(f,c)$. All details on this experiment are shown in \ref{res:stoch}. These unmistakenly assess the consistency of the shuffling methodology, allowing us to introduce the next experiment.

\subsection{Instances per class influence on the threshold}\label{exp:ic}
The main hypothesis of this paper is that there is a strong relation between optimal thresholds and target task $\hat{I_c}$, particularly the logarithmic reciprocal function formalised in (\ref{eq:log_recip}). All the results of the following experiments are shown in \ref{res:ic}.

Firstly, to study the influence of $\hat{I_c}$ alone, we fit a regression over a single target task: \textit{mit67TRTE}, and pre-trained model \textit{VGG16IN}. This task is balanced in $I_c$, and we obtain several thresholds by using stratified subsets of the task, corresponding to values $I_c$ multiples of 10 ([10,100]) (Figure \ref{fig:mit67reg}). To evaluate the goodness-of-fit, we use \textit{leave-one-out cross-validation} $R^2$ coefficient on these samples. This evaluation method is used with all regressions described in this section.

Secondly, to ensure generalisation of the regression, we fit a regression over the whole set of target tasks on \textit{VGG16IN}, \textit{VGG19IN} and \textit{VGG16P2} separately (Figure \ref{fig:sourcemodelregs}). In addition, we compare whether these three pre-trained models behave differently, as it may hint to the importance of the model's discriminativity.

Thirdly, we fit a regression on the target tasks that have low class imbalance, using model \textit{VGG16IN} (Figure \ref{fig:16prima}). We hypothesise that high imbalance behaves differently from the rest, as seen in Section \ref{res:stoch}. Finally, we evaluate this regression with the empirical thresholds of the subsets of \textit{mit67TRTE}, so as to observe if our fitted regression does generalise correctly for newer instances (Figure \ref{fig:primamit}).

\subsection{Practical impact of proposed model}
All previous experiments are directed at finding better thresholds for feature discriminativity assessment, and to do so through a trained and reliable model. To assess the impact of such model, we explore its effect on the $D_{KS}$ classified according to these thresholds. Since thresholds determine which feature-class pairs get discretised to either 0, 1 or -1, we perform this evaluation by measuring how many features change value by using the thresholds found by our model. The results are presented in Section \ref{res:err}.

\section{Results}

\subsection{Inner task stochasticity}\label{res:stoch}
Table \ref{tab:innerdatastoch} shows the threshold values and their standard deviation for each of the selected tasks. Notice all standard deviations are at least 2 orders of magnitude below the thresholds. This fact speaks for the consistency of the methodology.

\begin{table}[t]
    \caption{Threshold statistics for 4 target tasks, ordered by average instances per class descending}
    \label{tab:innerdatastoch}
    \centering
    \scalebox{1}{
    \begin{tabular}{lrrrrrr}
        \toprule
        Target task     & $t^-$ Avg  & $t^- \sigma$ & $t^+$ Avg & $t^+ \sigma$ & $I_c \pm \sigma$ \\
        \midrule
        mit67TRTE       & -0.109 & 0.00125 & 0.119 & 0.00050 & 100\\
        caltech101TRTE          & -0.140 & 0.00077 & 0.160 & 0.00030 & 89.66 $\pm$ 123.07 \\
        cub200TR      & -0.174 & 0.00112 & 0.195 & 0.00090 & 29.97 $\pm$ 0.17\\
        flowers102TR  & -0.284 & 0.00234 & 0.321 & 0.00238 & 10\\ 
        \bottomrule
    \end{tabular}
    }
\end{table}

\subsection{Instances per class influence on the threshold}\label{res:ic}

\begin{figure}
    \centering
    \hspace{-15pt}
    \begin{minipage}[t]{.5\textwidth}
        \centering
        \includegraphics[width=\linewidth]{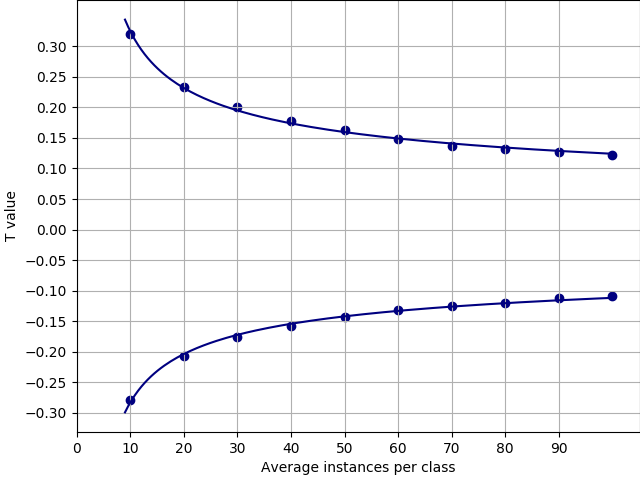}
        \vspace{-20pt}
        \caption{Regression of thresholds for subsets of mit67, with different number of instances per class. The dots correspond to empirical threshold values.}
        \label{fig:mit67reg}
    \end{minipage}
    \hspace{5pt}
    \begin{minipage}[t]{.45\textwidth}
        \centering
        \vspace{-140pt}
        \captionof{table}{$R^2$ values of each regression. \textit{bal.} stands for balanced regression.}
        \label{tab:r2}
        \vspace{16pt}
        \begin{tabular}{lrrrrrr}
        \toprule
        Experiment     & $t^-$  & $t^+$\\
        \midrule
        mit67 subsets & 0.986 & 0.990 \\
        VGG16IN & 0.944 & 0.962 \\
        VGG19IN & 0.920 & 0.935 \\
        VGG16P2 & 0.936 & 0.950 \\
        VGG16IN \textit{bal.} & 0.995 & 0.997 \\
        mit67 \textit{bal.} & 0.993 & 0.995 \\
        \bottomrule
    \end{tabular}
    \end{minipage}
\end{figure}

Figure \ref{fig:mit67reg} corresponds to the regression performed on the subsets of \textit{mit67TRTE}. We observe a surprisingly high $R^2$ coefficient for both positive and negative thresholds. This supports our claim that the thresholds are predictable from $I_c$.

\begin{figure}[htbp!]
\centering
\begin{subfigure}[htbp!]{0.47\textwidth}
\centering
    \includegraphics[width=\textwidth]{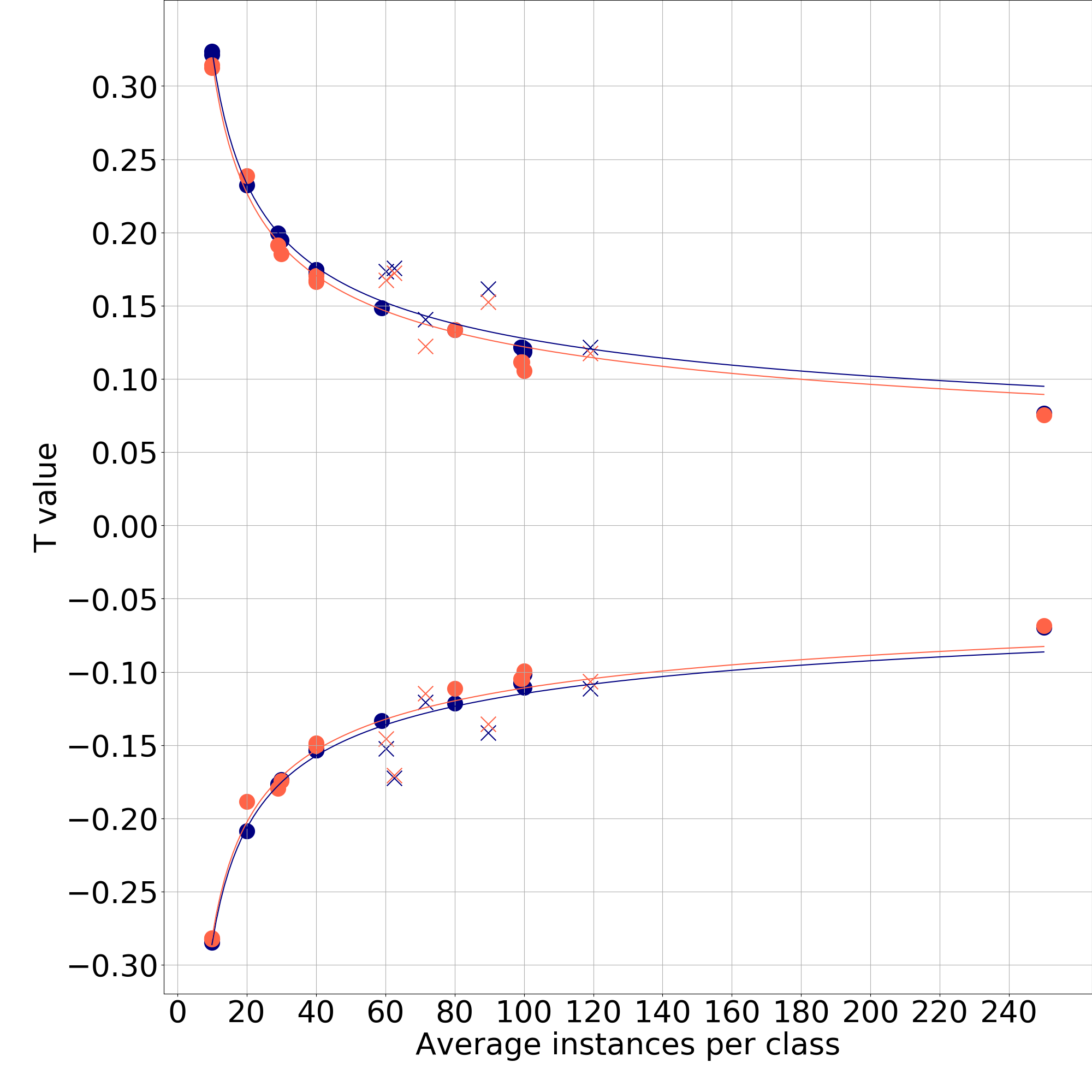}
    \caption{\textit{VGG16IN} (blue) versus \textit{VGG16P2} (orange).}
    \label{fig:imgpl2reg}
\end{subfigure}
\hfill
\begin{subfigure}[htbp!]{0.47\textwidth}
\centering
    \includegraphics[width=\textwidth]{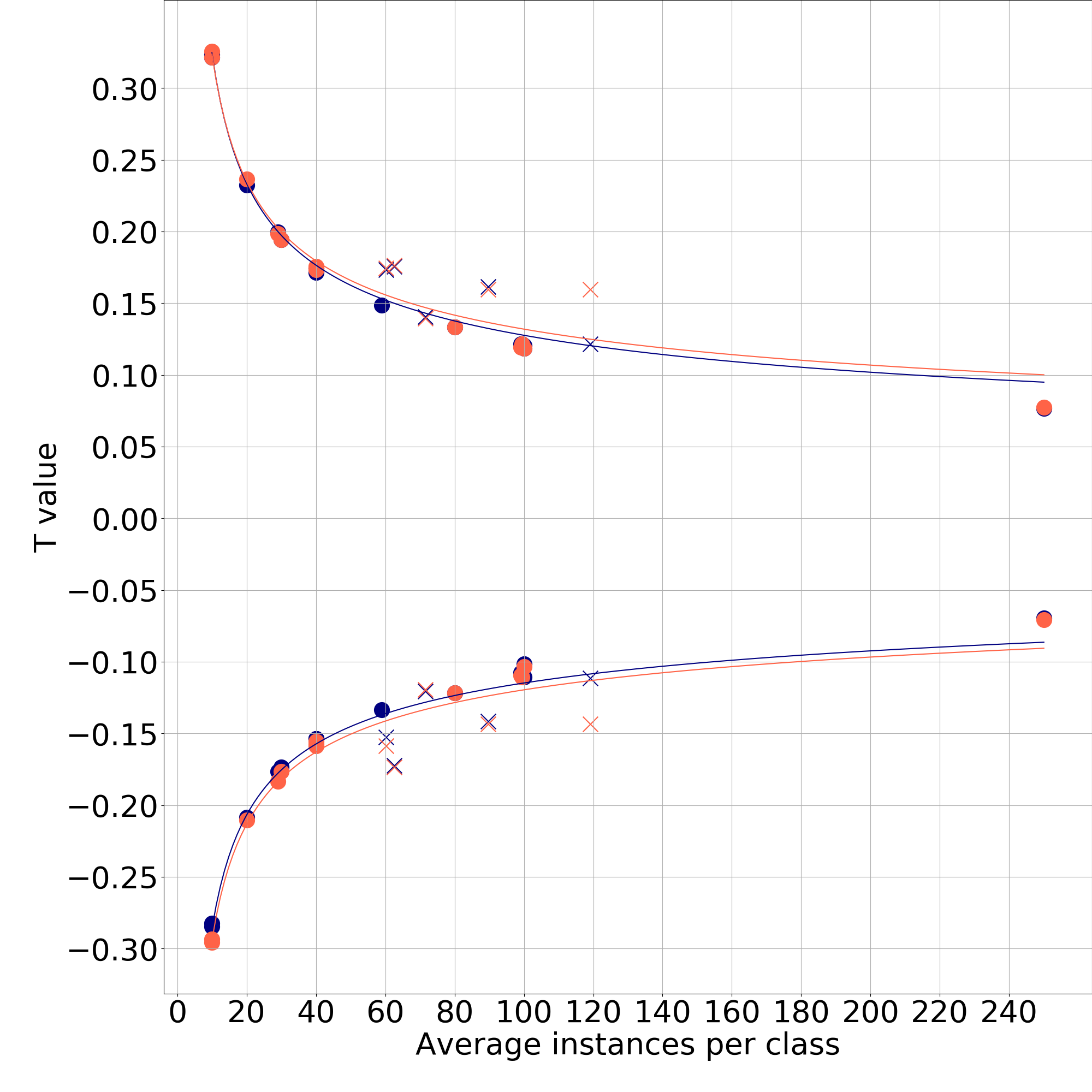}
    \caption{\textit{VGG16IN} (blue) versus \textit{VGG19IN} (orange).}
    \label{fig:1619reg}
\end{subfigure}
\caption{Regression over all target tasks and different pre-trained models.  Marked with X are the empirical threshold values of task partitions with $\sigma$ label distribution above 20.}
\label{fig:sourcemodelregs}
\end{figure}

\begin{figure}[htbp!]
\centering
\begin{subfigure}[htbp!]{0.47\textwidth}
\centering
    \includegraphics[width=\textwidth]{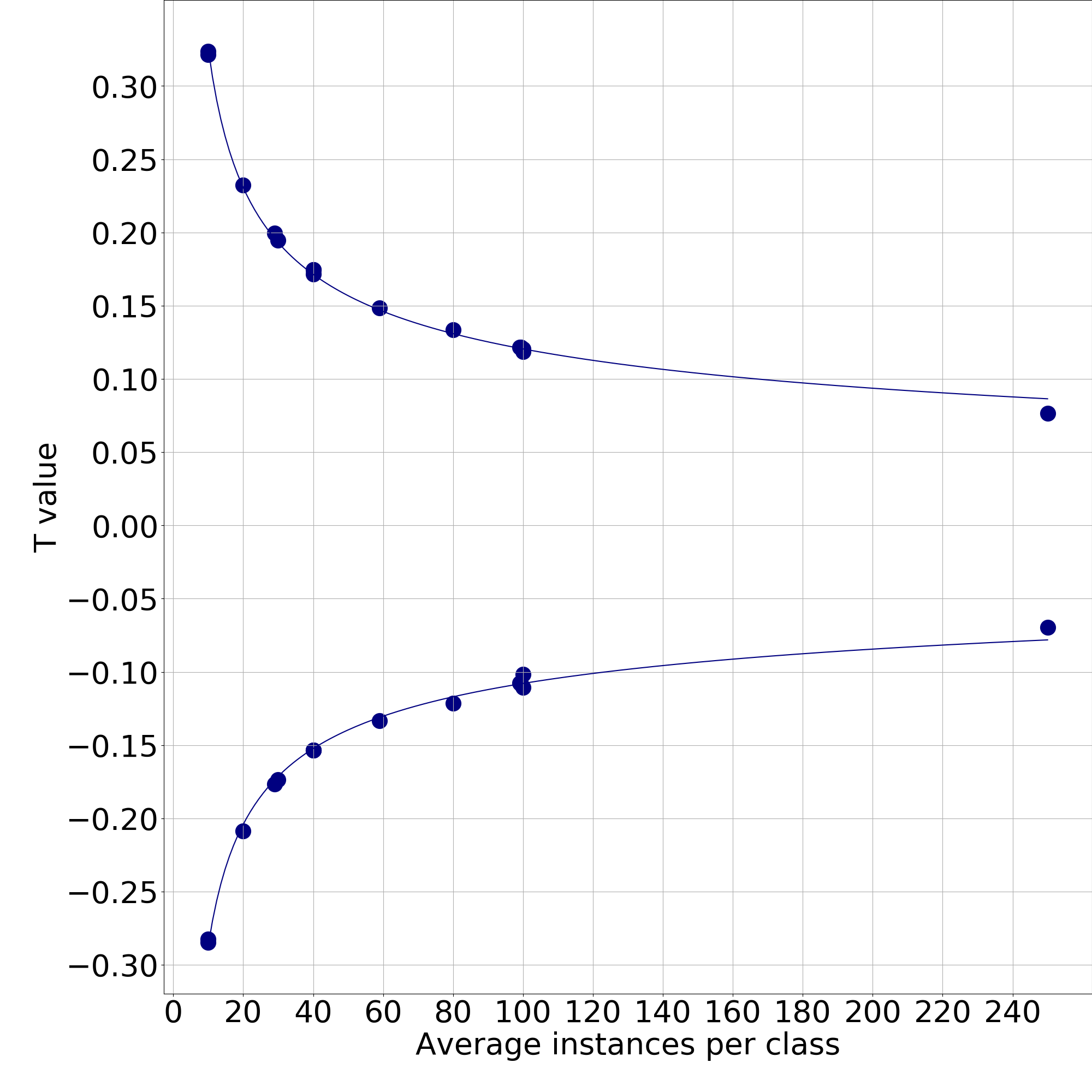}
    \caption{Regression with dots corresponding to empirical threshold values of the target tasks.}
    \label{fig:16prima}
\end{subfigure}
\hfill
\begin{subfigure}[htbp!]{0.47\textwidth}
\centering
\vspace{-10pt}
    \includegraphics[width=\textwidth]{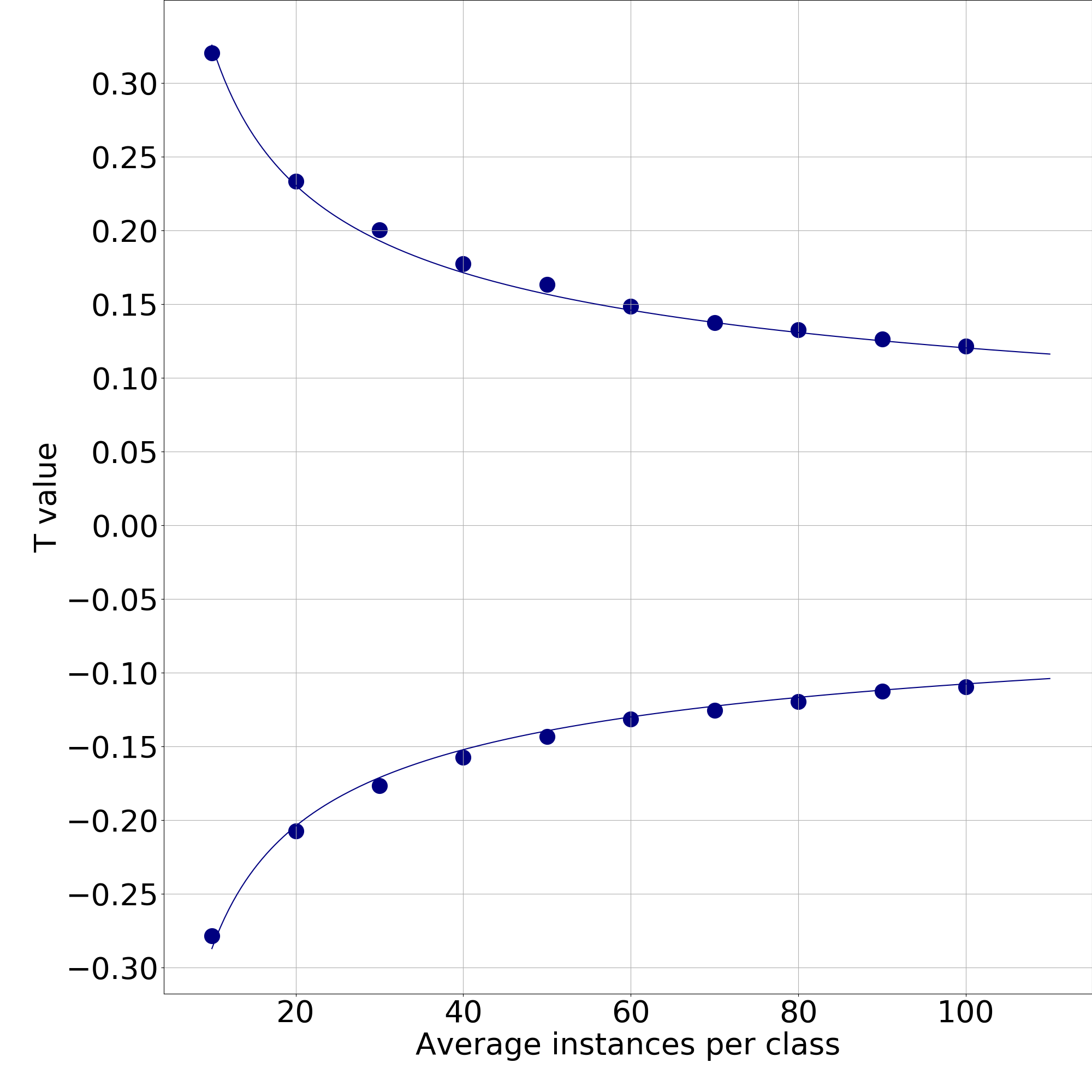}
    \caption{Regression with dots corresponding to \textit{mit67TRTE} cut to different $I_c$.}
    \label{fig:primamit}
\end{subfigure}
\caption{Balanced regression.}
\end{figure}

In Figure \ref{fig:sourcemodelregs} we expose the difference in behaviour caused by altering the properties of pre-trained models. In Figure \ref{fig:imgpl2reg}, we compare \textit{VGG16IN} and \textit{VGG16P2} (different source tasks). In Figure \ref{fig:1619reg} we compare \textit{VGG16IN} and \textit{VGG19IN} (different architectures). In both plots of Figure \ref{fig:sourcemodelregs} we observe a consistent set of outliers that are not as well adjusted as the others. Remarkably, these correspond to tasks with significant class imbalance (standard deviation above 20, as seen in Table \ref{tab:datasets}). For clarity, these data points have been marked with $x$ in the previous figures. Figure \ref{fig:16prima} is a regression on \textit{VGG16IN} having removed these tasks: \textit{stanforddogsTE}, \textit{woodTR}, \textit{flowers102TE}, \textit{caltech101TRTE}, \textit{caltech256TRTE}. We refer to this as the \textit{balanced regression}. Figure \ref{fig:primamit} shows the previously fitted balanced regression, on top of the threshold values from the subsets of \textit{mit67TRTE}.

The $R^2$ values of all these regressions are presented in Table \ref{tab:r2}.

\subsection{Imbalance error and influence}\label{res:err}


To evaluate the impact of our methodology, we perform a study on the difference between the original thresholds for \textit{VGG16IN} obtained with the stochastic method, and the predicted with the regression on \textit{VGG16IN} with no filtering. In Table \ref{tab:thresholdinf} we record the threshold values as well as the percentage of changes. Coherently, the ones with higher amount of changes are the imbalanced tasks, as well as the \textit{food101TE} (this particular case is discussed in Section \ref{discussion}).

\section{Discussion}\label{discussion}
One of our initial hypothesis was that the standard deviation due to the stochasticity is small compared to the values obtained. This seems validated by the results in Table \ref{tab:innerdatastoch} as the standard deviation is two orders of magnitude smaller than the mean threshold value. This table also indicates the presence of an inverse correlation between standard deviation for the thresholds and $\hat{I_c}$ in the balanced tasks. A more complete analysis would be needed to further validate this point.

Regarding the results from Table \ref{tab:r2}, we observe that the thresholds are highly predictable from $I_c$. We can attribute part of the error to neglecting class imbalance; notice how by removing the imbalanced tasks we drastically raise the $R^2$. This extraordinarily high predictability hints to the existence a mathematical relationship. Another significant finding is that the balanced regression (fitted with all balanced tasks) characterises better the \textit{mit67TRTE} subsets' thresholds than the regression tailored for them. We attribute this to the sample size.

Comparing between pre-trained models (Figure \ref{fig:sourcemodelregs}) we find that regressions are almost superposed. We hypothesise the difference comes from a different discriminativity across the pre-trained models \wrt the targets. Even though it seems that a different topology (Figure \ref{fig:1619reg}) yields a greater difference than different source task (Figure \ref{fig:imgpl2reg}), this is actually due to the outlier \textit{caltech256TRTE}. If this task is removed, the difference is much less than that between the source tasks. The impact of both factors (source task and architecture) is thus minimal.

\begin{table}[htbp!]
    \caption{Threshold influence}
    \label{tab:thresholdinf}
    \centering
    \begin{tabular}{lrrrrrr}
        \toprule
        & $t^-$ & & $t^+$ & & Percentage of \\
        Task     & original  & predicted & original & predicted &   group changes\\
        \midrule
caltech101TRTE & -0.1415 & -0.1182 & 0.1615 & 0.1317 & 7.691 \\
caltech256TRTE & -0.1115 & -0.1077 & 0.1215 & 0.1196 & 1.016 \\
catsdogsTR & -0.1085 & -0.1142 & 0.1215 & 0.1271 & 2.105 \\
catsdogsTE & -0.1075 & -0.1143 & 0.1215 & 0.1272 & 2.320 \\
catsdogsTRTE & -0.0825 & -0.0915 & 0.0925 & 0.1011 & 3.371 \\
cub200TR & -0.1735 & -0.1753 & 0.1945 & 0.1972 & 0.720 \\
cub200TE & -0.1765 & -0.1776 & 0.1995 & 0.1999 & 0.276 \\
cub200TRTE & -0.1335 & -0.1364 & 0.1485 & 0.1526 & 1.241 \\
flowers102TR & -0.2825 & -0.2870 & 0.3215 & 0.3254 & 0.867 \\
flowers102VAL & -0.2845 & -0.2870 & 0.3235 & 0.3254 & 0.439 \\
flowers102TE & -0.1525 & -0.1353 & 0.1735 & 0.1514 & 5.503 \\
food101TE & -0.0695 & -0.0853 & 0.0765 & 0.0939 & 8.668 \\
mit67TR & -0.1215 & -0.1228 & 0.1335 & 0.1370 & 0.848 \\
mit67TE & -0.2085 & -0.2069 & 0.2325 & 0.2335 & 0.445 \\
mit67TRTE & -0.1105 & -0.1140 & 0.1205 & 0.1269 & 1.850 \\
stanforddogsTR & -0.1015 & -0.1140 & 0.1185 & 0.1269 & 4.553 \\
stanforddogsTE & -0.1205 & -0.1276 & 0.1405 & 0.1425 & 2.098 \\
texturesTR & -0.1535 & -0.1570 & 0.1745 & 0.1762 & 0.969 \\
texturesVAL & -0.1535 & -0.1570 & 0.1715 & 0.1762 & 1.473 \\
texturesTE & -0.1535 & -0.1570 & 0.1745 & 0.1762 & 0.957 \\
woodTR & -0.1725 & -0.1336 & 0.1755 & 0.1494 & 10.025 \\

        \bottomrule
    \end{tabular}
\end{table}

We find an outlier in the balanced regression (Figure \ref{fig:16prima} and Table \ref{tab:thresholdinf}): \textit{food101TE}. While the task is balanced, the data point is the furthest away from the line, and has the second highest percentage of changes. We think this is caused by this target task being less discriminated against. Unlike other tasks where the average absolute $D_{KS}$ is above $0.2$, for this one is $0.15$ (near the value of our predicted threshold). This means that feature-class pairs are not very discriminative. To optimise the threshold, the original method lowers the absolute value of the thresholds, raising the amount of noise but also of information. Since there are many feature-class pairs in this interval, small movements of the threshold heavily influence the amount of changes.

\section{Conclusions and Future Work}

The purpose of this paper was to find a suitable yet simple method to determine thresholds for discriminative noise trade-off in feature extraction. We outline our conclusions next. 

\begin{enumerate}
    \item The stochasticity of shuffling does not heavily modify the thresholds, it only slightly deforms the $D'_{KS}$ distribution (Section \ref{res:stoch}). 
    \item The number of instances per class can be transformed into the final threshold with the formula \ref{eq:log_recip}, obtaining a reduced error (Sections \ref{res:stoch}, \ref{res:err}). 
    \item Most of the previous error has been shown to come mostly from the class imbalance (Section \ref{res:ic}).
    \item The small remaining error might come from the difference between source and target tasks (\textit{food101TE} in Sections \ref{res:ic} \ref{res:err}).
\end{enumerate}

In this work, we identified potential improvements for future work.

\begin{enumerate}
    \item The imbalance of the dataset's classes produces an error which we believe could be integrated in the regression function.
    \item The difference between VGG topologies seems to be much smaller than between source tasks when considering a pre-trained models. We have yet to see if this applies to the rest of CNN topologies.
    \item Information about the real discriminativity, such as the mean absolute discriminativity, might reduce the number of changed features. This would reduce the error in datasets such as \textit{food101}.
\end{enumerate}

\section*{Acknowledgements}
This work is partially supported by BSC-IBM Deep Learning Center agreement, the Spanish Government through Programa Severo Ochoa (SEV-2015-0493), the Spanish Ministry of Science and Technology through TIN2015-65316-P project and the Generalitat de Catalunya (contract 2017-SGR-1414).
\bibliographystyle{ieeetr}
\bibliography{biblio}

\end{document}